%% file: acl_latex.tex
\pdfoutput=1

\documentclass[11pt]{article}

\usepackage[final]{acl}

\usepackage{times}
\usepackage{latexsym}

\usepackage[T1]{fontenc}

\usepackage[utf8]{inputenc}

\usepackage{microtype}

\usepackage{graphicx}
\usepackage{svg}      
\usepackage{booktabs}
\usepackage{multirow}
\usepackage{float}
\usepackage{listings}
\usepackage{multicol}
\usepackage{arydshln}

\lstdefinestyle{bash}{
  language=bash,
  backgroundcolor=\color{gray!10},
  basicstyle=\ttfamily\footnotesize,
  breaklines=true,
  frame=single,
  columns=fullflexible
}

\lstdefinestyle{promptstyle}{
    basicstyle=\ttfamily\small,
    backgroundcolor=\color{gray!10},
    frame=single,
    breaklines=true,
    breakatwhitespace=true,
    columns=fullflexible,
    captionpos=b
}

\title{Structuring Radiology Reports: \\Challenging LLMs with Lightweight Models}

\def\huggingface{\raisebox{-1.5pt}{\includegraphics[height=1.05em]{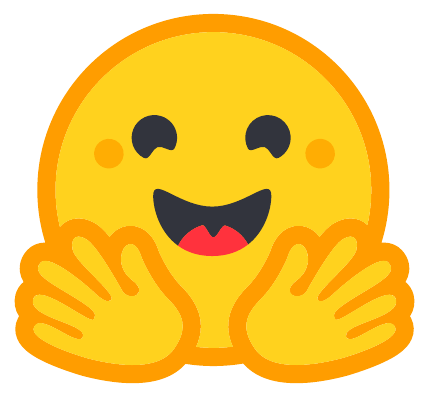}}}
\def\github{\raisebox{-1.5pt}{\includegraphics[height=1.05em]{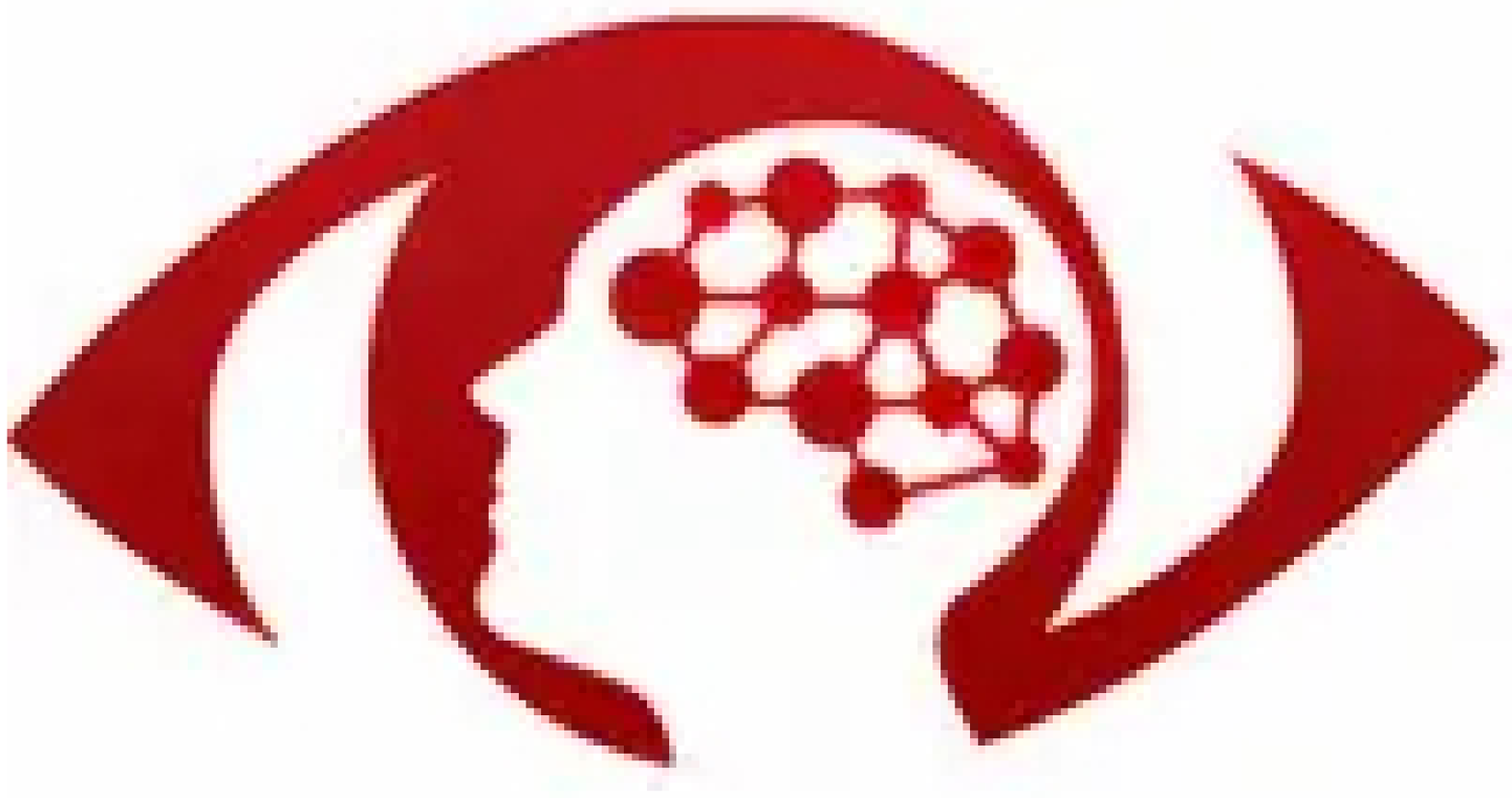}}}

\newcommand{\hflink}{https://huggingface.co/StanfordAIMI}
\newcommand{\ghlink}{https://stanford-aimi.github.io/structuring.html}
\author{
  \textbf{Johannes Moll\textsuperscript{1,2}},
  \textbf{Louisa Fay\textsuperscript{1}},
  \textbf{Asfandyar Azhar\textsuperscript{1,3}},
  \textbf{Sophie Ostmeier\textsuperscript{1}},
\\
  \textbf{Tim Lueth\textsuperscript{2}},
  \textbf{Sergios Gatidis\textsuperscript{1}},
  \textbf{Curtis P. Langlotz\textsuperscript{1}},
  \textbf{Jean-Benoit Delbrouck\textsuperscript{1,4}}
\\
  \textsuperscript{1}Stanford University,
  \textsuperscript{2}Technical University of Munich,
  \\
  \textsuperscript{3}Carnegie Mellon University,
  \textsuperscript{4}HOPPR\\
  \small{
    \textbf{Correspondence:} \href{mailto:jomoll@stanford.edu}{jomoll@stanford.edu}
  }\\
  \begin{tabular}{rl}
        \huggingface & \url{\hflink}\\
        \github      & \url{\ghlink}
    \end{tabular}
}

\usepackage{comment}
\usepackage{tcolorbox}

\begin{document}
\maketitle

\begin{abstract}
Radiology reports are critical for clinical decision-making but often lack a standardized format, limiting both human interpretability and machine learning (ML) applications. While large language models (LLMs) have shown strong capabilities in reformatting clinical text, their high computational requirements, lack of transparency, and data privacy concerns hinder practical deployment. To address these challenges, we explore lightweight encoder-decoder models (<300M parameters)—specifically T5 and BERT2BERT—for structuring radiology reports from the MIMIC-CXR and CheXpert Plus datasets. We benchmark these models against eight open-source LLMs (1B–70B parameters), adapted using prefix prompting, in-context learning (ICL), and low-rank adaptation (LoRA) finetuning. Our best-performing lightweight model outperforms all LLMs adapted using prompt-based techniques on a human-annotated test set. While some LoRA-finetuned LLMs achieve modest gains over the lightweight model on the Findings section (BLEU 6.4\%, ROUGE-L 4.8\%, BERTScore 3.6\%, F1-RadGraph 1.1\%, GREEN 3.6\%, and F1-SRR-BERT 4.3\%), these improvements come at the cost of substantially greater computational resources. For example, LLaMA-3-70B incurred more than 400 times the inference time, cost, and carbon emissions compared to the lightweight model. These results underscore the potential of lightweight, task-specific models as sustainable and privacy-preserving solutions for structuring clinical text in resource-constrained healthcare settings.
\end{abstract}

\section{Introduction}
\input{1_Introduction}

\section{Related Work}
\input{2_RelatedWork}

\section{Methods}
\label{sec:methods}
\input{3_Methods}

\section{Results}
\label{sec:results}
\input{4_Results}

\section{Discussion}
\label{sec:5_discussion}
\input{5_Discussion}

\section{Conclusion}
\input{6_Conclusion}
\newpage
\section*{Limitations}

First, as discussed in Section~\ref{sec:met_data}, the labels used for training our specialized models and adapting the LLMs were generated from MIMIC-CXR and CheXpert Plus reports using GPT-4 as a weak annotator. While our prompt builds on previous work, we refined it to better align with our task’s requirements (e.g., explicitly specifying organ systems for the Findings section). However, GPT-4 may introduce biases, and to mitigate this, we evaluate model performance on an independent test set annotated by five radiologists.

\noindent Second, both MIMIC-CXR and CheXpert Plus originate from hospitals in the United States - Beth Israel Deaconess Medical Center (Boston, MA) and Stanford Hospital (Stanford, CA) - and contain only chest X-rays from adult patients. As a result, these datasets may lack demographic diversity, potentially limiting generalizability to other populations.

\noindent Third, as described in Section~\ref{sec:methods}, all models take full free-form reports as input and generate structured reports comprising the following sections: Exam Type, History, Technique, Comparison, Findings, and Impression. However, for quantitative evaluation, we focus exclusively on Findings and Impression, as these sections are clinically critical and exhibit the highest variability. Other sections, such as Exam Type and History, often remain unchanged and can be directly copied from the original report, making them less relevant for assessing model performance.

\noindent Fourth, 1-shot and 2-shot ICL examples were manually selected from the training set to best represent the data distribution. While we initially applied algorithmic methods to optimize alignment, manual selection proved to improve performance. This introduces a potential selection bias, which may affect the generalizability of our ICL results.

\noindent Fifth, while we initially experimented with full-parameter finetuning for select LLMs, we found that it did not yield substantial performance improvements over LoRA. Given the significantly higher computational and time demands of full finetuning, we opted to use LoRA as an efficient adaptation strategy for all LLMs within our resource constraints.

\noindent Sixth, we initially also evaluated GPT-4 using prefix prompting and ICL. However, since it was used for data annotation and provided as a reference for radiologist, its results may be biased in its favor. To account for this, we excluded GPT-4 from the discussion to avoid misleading comparisons.
  
\noindent Seventh, while we expected the LLMs—particularly the larger models—to outperform the lightweight model given their scale, this was not consistently observed under our current finetuning setup. Although we performed basic hyperparameter tuning and employed established adaptation techniques, the finetuning process may not have been sufficiently extensive or optimized to fully leverage the capabilities of these models. This is especially true for LLaMA-3-70B, which was limited to a single epoch of training due to computational constraints. 

\noindent Eighth, while our selection of LLMs aims to represent both the current state of the art and a range of model sizes, one could argue for the inclusion of more domain-specific models tailored to the medical field. We include MedAlpaca-7B as a representative example, but find that it underperforms compared to general-domain models of similar scale, suggesting that current medicine-specific LLMs may not yet offer a clear advantage for the structuring task evaluated here.

\section*{Acknowledgements}
This work was supported in part by the Medical Imaging and Data Resource Center (MIDRC), funded by the National Institute of Biomedical Imaging and Bioengineering (NIBIB) of the National Institutes of Health under contract 75N92020D00021 and through The Advanced Research Projects Agency for Health (ARPA-H).


\bibliography{literature}

\newpage
\onecolumn
\appendix
\section{Appendix}
\label{sec:appendix}
\input{Appendix}

\end{document}

%% file: 1_Introduction.tex
Radiology reports play a critical role in clinical workflows by summarizing imaging findings that guide medical decisions \cite{kahn2009toward}. However, variations in reporting style due to individual and institutional practices as well as regional guidelines create inconsistencies that hinder interpretability for physicians and patients \cite{hartung2020create}. Moreover, the lack of structured formats limits their usefulness as training data for machine learning (ML) applications \cite{dos2023esr, steinkamp2019toward}.

Large language models (LLMs) offer a promising solution for generating structured reports from free-form text \cite{adams2023leveraging, busch2024large, hasani2024evaluating}. However, deploying these models locally remains infeasible for most institutions due to the significant computational resources required \cite{zhang2025revolutionizing}. Cloud-based solutions provide an alternative but introduce concerns related to data security, confidentiality, and regulatory compliance \cite{arshad2023chatgpt, thirunavukarasu2023large}. While proprietary LLMs can also be accessed via Application Programming Interface (API), this approach entails drawbacks such as dependency on a third-party vendor, potential cost increases and unpredictable changes in usage terms \cite{tian2024opportunities}. These limitations highlight the need for smaller, open-source models that can be deployed on-device with minimal hardware requirements.

\begin{figure*}
    \centering
    \includegraphics[width=\linewidth]{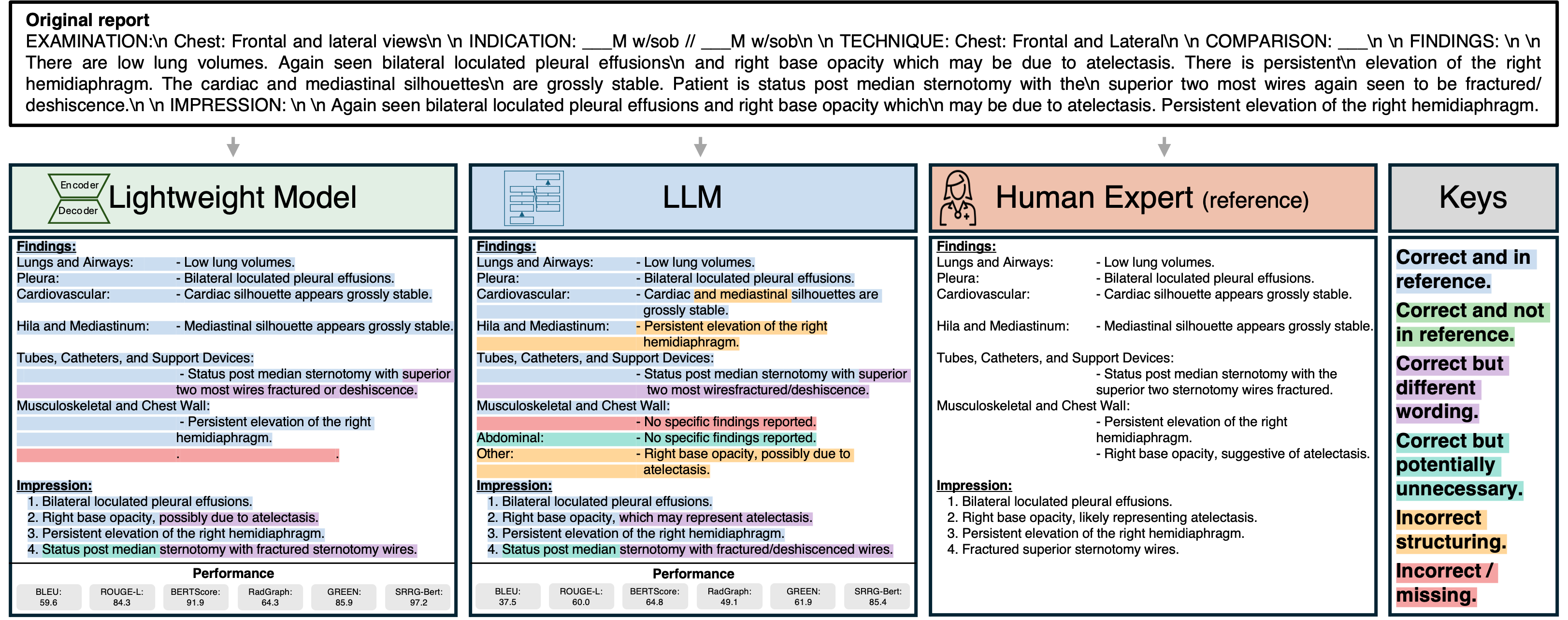}
    \caption{Overview of our study and qualitative comparison. An unstructured radiology report is structured using lightweight, task-specific models and adapted large language models (LLMs) compared to human expert annotations.}
    \label{fig:qualitative}
\end{figure*}

To address these challenges, we propose lightweight (<300M parameters), task-specific models for structuring free-text chest X-ray radiology reports (see Figure~\ref{fig:qualitative}) efficiently. These models substantially reduce computational demands \cite{chen2024improving}, eliminating the need for cloud-based hosting, and enhancing data security by enabling offline deployment. We train these models on the MIMIC-CXR \cite{johnson2019mimic} and CheXpert Plus \cite{chambon2024chexpert} datasets and structure the originally free-form reports with GPT-4~\cite{achiam2023gpt} as a weak annotator, enabling large-scale supervision. We evaluate model performance on an independent test set, annotated by five radiologists \cite{anonymous2025structured}. 
Our contributions include:

\begin{itemize} 
    \item \textbf{Lightweight Model Development and Evaluation}: We train and systematically evaluate lightweight (<300M parameters), task-specific T5 and BERT2BERT models for the task of structuring radiology reports. 
    \item \textbf{Analysis of LLMs and Adaptation Techniques}: We assess the performance of five LLMs (3-8B parameters) under different adaptation strategies (prefix prompting, in-context learning (ICL), low-rank adaptation (LoRA)).
    \item \textbf{Benchmarking and Cost Analysis}: We benchmark lightweight models against LLMs of increasing size, considering model performance on the BLEU, ROUGE-L, BERTScore, F1-RadGraph, GREEN, and F1-SRRG-Bert metrics, as well as training time, inference speed and costs, and environmental impact.\end{itemize}

%% file: 2_RelatedWork.tex
\begin{figure*}[t] 
    \centering
    \includegraphics[width=\linewidth]{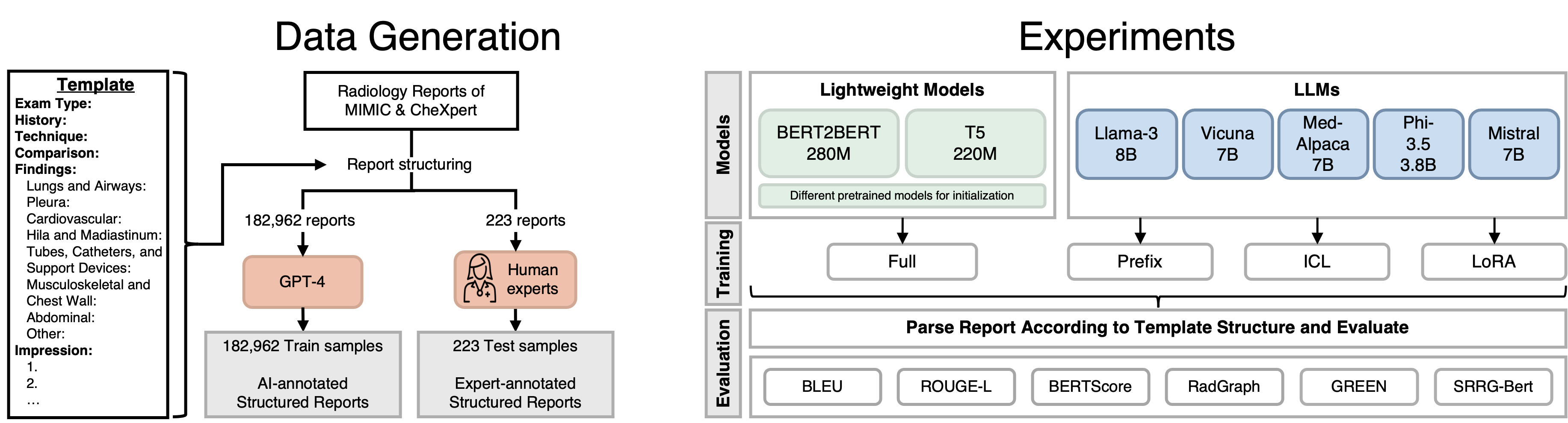}
    \caption{Left: Dataset generation from free-form radiology reports to structured radiology reports using GPT-4 (AI-based) and human experts (manual annotation). Right: Overview of our experiments including selection of lightweight models and LLMs, training/adaptation methods, and evaluation strategy and metrics.}
    \label{fig:data_training}
\end{figure*}
\noindent \textbf{Beyond LLMs: Lightweight Models for Medical Text Processing}\\
Recent studies have explored the use of LLMs, namely GPT-3.5 \cite{openai_gpt35} and GPT-4, to transform free-form radiology reports into structured formats \cite{adams2023leveraging, bergomi2024reshaping, hasani2024evaluating}. A recent review by Busch et al. highlights that these approaches achieve low error rates and minimal accuracy loss compared to human experts \cite{busch2024large}. However, their reliance on proprietary architectures, lack of transparency, and restrictions on patient data privacy pose significant challenges for clinical deployment \cite{khullar2024large, rezaeikhonakdar2023ai}. 
To address these limitations, similar tasks in medical NLP have adopted lightweight, task-specific models that maintain high accuracy while considerably reducing computational costs \cite{chen2024improving, griewing2024proof, pecher2024comparing}. Existing task-specific models for radiology NLP fall into two categories: hybrid models and lightweight transformer models. Hybrid models combine rule-based methods with deep learning, enforcing domain-specific constraints but lacking flexibility \cite{gabud2023hybrid}. In contrast, lightweight transformer models have been successfully applied to relation extraction, report coding, and summarization \cite{jain2021radgraph, yan2022radbert, van2023radadapt}. While they require careful tuning to avoid hallucinations and overfitting, recent studies suggest that well-tuned lightweight models can match larger LLMs in accuracy while being far more computationally efficient \cite{pecher2024comparing}. Our work builds on this foundation by introducing a lightweight, task-specific model explicitly optimized for structured radiology report generation. 

\noindent \textbf{Model Adaptation and Finetuning}\\
Prior work has explored a range of adaptation strategies for LLMs, from prompt-based methods to parameter-efficient finetuning (PEFT) and full finetuning, each balancing performance, data requirements, and computational cost. Prompting techniques such as prefix prompting and ICL \cite{brown2020language,lampinen2022can} adapt models without modifying their weights. Prefix prompting typically provides instructions to guide model responses, while ICL enhances adaptation by incorporating task-specific examples within the prompt. However, these methods suffer from context length constraints and sensitivity to prompt phrasing \cite{li2023long}. PEFT techniques like LoRA \cite{hu2021lora}, prefix-tuning \cite{li2021prefix}, and adapter layers \cite{houlsby2019parameter} enable efficient adaptation with minimal computational overhead, making them well-suited for clinical NLP. While effective in low-data settings, PEFT often struggles with complex reasoning and generalization across domains \cite{lialin2023scaling}. In contrast, full finetuning updates all model parameters, often achieving stronger adaptation when sufficient labeled data and computational resources are available. Building on this, our approach applies full finetuning to lightweight models while leveraging GPT-4-generated structured labels to address data scarcity, enabling large-scale supervised training while preserving domain-specific accuracy.

\noindent \textbf{AI-Based Dataset Generation}\\
A major challenge in developing models for structuring radiology reports is the limited availability of high-quality annotated datasets, i.e., datasets that contain both free-form and corresponding structured reports. Recent work in similar fields has explored leveraging LLMs such as GPT-4 as weak annotators to generate labels, providing a scalable alternative to manual annotation~\cite{liyanage2024gpt, savelka2023can}. Despite their successes, studies suggest that models trained on GPT-generated data should still be rigorously evaluated against human-annotated ground truth to ensure reliability and validity~\cite{pangakis2023automated}.

%% file: 3_Methods.tex
In this study, we transform free-text chest X-ray radiology reports into a standardized format using deep learning. The structured reports follow a predefined template based on 'RPT144' of RSNA’s RadReport Template Library \cite{rsna_radreport}. This template comprises the sections: Exam Type, History, Technique, Comparison, Findings, and Impression. The Findings section is further organized into organ systems: 'Lungs and Airways', 'Pleura', 'Cardiovascular', 'Tubes, Catheters, and Support Devices', 'Musculoskeletal and Chest Wall', 'Abdominal', and 'Other'. The Impression section is structured as a numbered list, prioritizing the most clinically relevant findings. As shown in Figure~\ref{fig:data_training}, this template is incorporated into the prompt during data annotation, and deviations from it in a structured report are penalized during evaluation. Unlike previous approaches that rely on large, general-purpose models like GPT-4, we explore the effectiveness of lightweight, task-specific models for this task.

\subsection{Data}
\label{sec:met_data}
We use unstructured radiology reports from the publicly available MIMIC-CXR \cite{johnson2019mimic} and CheXpert Plus \cite{chambon2024chexpert} datasets, preserving their original training and validation splits. To train our models in a supervised manner, we employed GPT-4 as a weak annotator, using the prompt provided in Appendix~\ref{sec:A_prompt} to generate structured reports that conform to our template. We obtained a total of 182,962 reports, 125,447 samples from MIMIC-CXR and 57,515 from CheXpert Plus. For evaluation and benchmarking, we conducted a human expert review of 223 reports, comprising 161 from the MIMIC-CXR test set and 72 from the CheXpert Plus validation set. Five board-certified radiologists from our institution reviewed the structured reports alongside their original free-form counterparts, assessing them for errors and adherence to our predefined template (detailed in \cite{anonymous2025structured}).

\subsection{Evaluation Strategies}
\label{sec:metrics}
Even though all models generate full reports, we focus our quantitative analysis on the Findings and Impression sections due to their clinical significance. Before applying our metrics, we parse these sections to assess adherence to the predefined template. In the Findings section, we identify predefined organ system headers (e.g., 'Lungs and Airways', 'Cardiovascular') and extract their corresponding observations. Metrics are computed separately for each organ system and then averaged across all identified systems. In the Impression section, we enforce a sequentially numbered format and flag any inconsistencies in ordering. To assess both linguistic quality and clinical accuracy, we use a combination of lexical and radiology-specific metrics.\\
\noindent \textbf{Lexical Metrics} To ensure comprehensive evaluation of text quality, we apply the following metrics: 
\textit{BLEU} \cite{papineni2002bleu} measures n-gram overlap, serving as a proxy for fluency and syntactic similarity. 
\textit{ROUGE-L} \cite{lin2004rouge} evaluates the longest common subsequence, capturing sentence-level similarity. 
\textit{BERTScore} \cite{zhang2019bertscore}  computes semantic similarity by comparing contextual embeddings from a pretrained transformer model.
\begin{figure*}
    \centering
    \includegraphics[width=\linewidth]{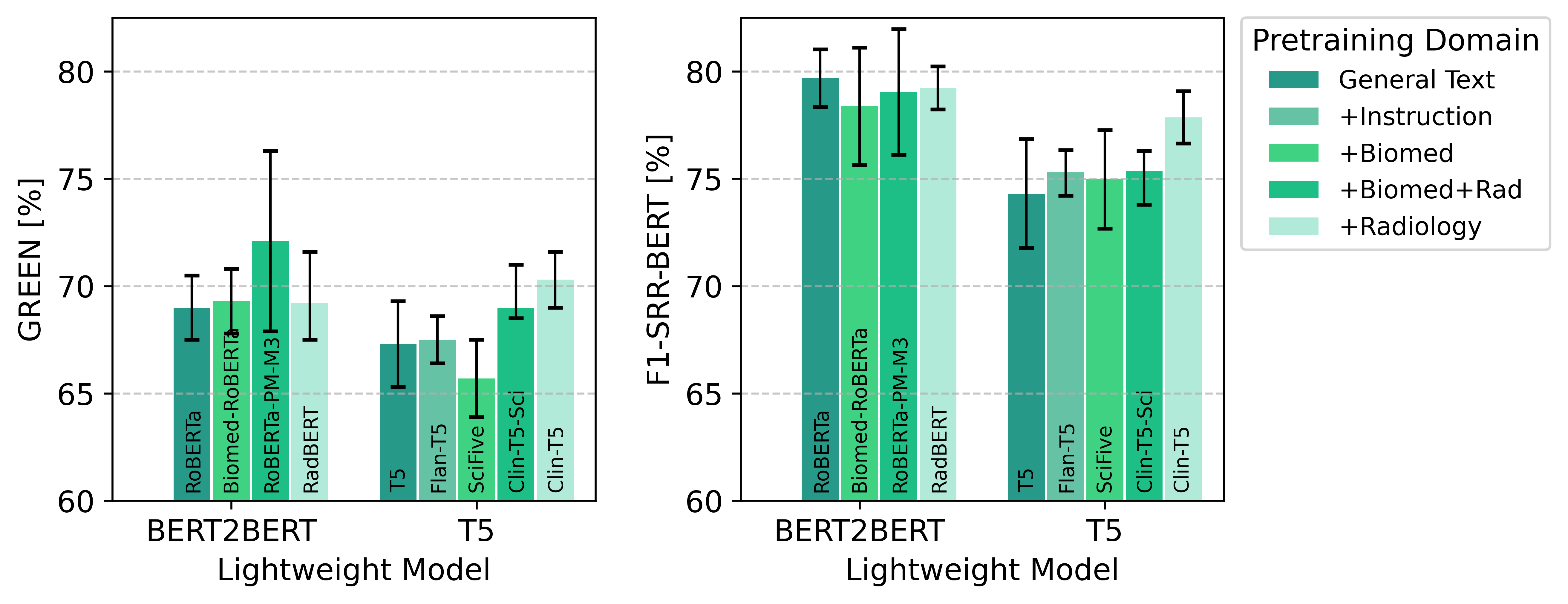}
    \caption{Performance comparison of lightweight models, initialized from pretrained models of increasing domain relevance. The plot shows the finetuned BERT2BERT and T5 models evaluated using GREEN (left) and F1-SRR-BERT (right), initialized from various pretrained models, with pretraining datasets ranging from general text (least domain-specific) to radiology (most domain-specific). Error bars denote $95\%$ confidence intervals over the three training runs.}
    \label{fig:t5_domain}
\end{figure*}\\
\noindent \textbf{Radiology-Specific Metrics} To capture clinical accuracy, we apply the following metrics: 
\textit{F1-RadGraph} \cite{delbrouck2022improving,yu2023evaluating} evaluates the precision and recall of key clinical terms and relationships extracted from generated reports. 
\textit{GREEN} \cite{ostmeier2024green} assesses the factual correctness of generated radiology reports using a finetuned LLM. 
\textit{F1-SRRG-Bert} \cite{anonymous2025structured} uses a fine-tuned BERT model to classify extracted findings into 55 disease labels, assigning each as Present, Absent, or Uncertain. It then computes the F1-score by comparing predictions from the generated report to the ground truth.\\ 
Throughout this paper, our visualizations primarily focus on GREEN and F1-SRR-BERT, as GREEN correlates most strongly with expert evaluations of clinical accuracy~\cite{ostmeier2024green}, while F1-SRR-BERT was specifically developed for the task of structured reporting, making their combination effective for assessing structured radiology reports.

\subsection{Lightweight Models}
\label{sec:met_lightweight}
We introduce lightweight models, which are specifically trained to structure radiology reports according to a predefined template. Our lightweight models are based on encoder-decoder architectures given their recent success in similar tasks such as radiology report generation \cite{aksoy2023radiology, chen2024act} and radiology report summarization \cite{de2024leveraging,van2023radadapt, zhang2018learning}. Specifically, we focused on two architectures, \textit{T5-Base} \cite{raffel2020exploring}, which has 223M parameters, and \textit{BERT2BERT} \cite{rothe2020leveraging}, where two identical BERT models are used as the encoder and decoder, resulting in a total of 278M parameters. To investigate the influence of pretraining domains, we initialize our models with the parameters from five open-source T5 variants (Table~\ref{tab:t5_models}) - \textit{T5-Base} \cite{raffel2020exploring}(general text), \textit{Flan-T5-Base} \cite{chung2024scaling}(instruction-tuning), \textit{SciFive} \cite{phan2021scifive}(biomedical text), \textit{Clin-T5-Sci} \cite{lehman2023clinical}(biomedical text and radiology reports), and \textit{Clin-T5-Base} \cite{lehman2023clinical}(radiology reports) - and four BERT variants (Table~\ref{tab:roberta_models}) - \textit{RoBERTa-base} \cite{liu2019roberta}(general text), \textit{BioMed-RoBERTa} \cite{gururangan2020don}(biomedical text), \textit{RoBERTa-base-PM-M3-Voc-distill-align} \cite{lewis2020pretrained}(for simplicity named RoBERTa-PM-M3 here, biomedical text and radiology reports), and \textit{RadBERT-RoBERTa} \cite{yan2022radbert}(radiology reports). We train our lightweight models end-to-end, updating all parameters, for a maximum of ten epochs using a cosine learning rate scheduler with an initial learning rate of $1e^{-4}$, an effective batch size of $128$, and the Adam optimizer. A detailed description of hyperparameters can be found in Appendix~\ref{sec:A_endtoend}. To account for variability, each configuration is trained three times with different random seeds. Following prior work \cite{van2023radadapt}, we rank pretraining datasets by relevance, assuming radiology reports to be the most relevant, followed by biomedical text (e.g., PubMed abstracts) and general-domain text (e.g., Wikipedia). However, we acknowledge that this ranking is inherently subjective and may vary depending on the specific task.

\subsection{Comparison LLMs}
\label{sec:met_llm}
To benchmark our lightweight models (<300M parameters), we first conduct a comprehensive comparison with instruction-tuned LLMs ranging from 3 to 8 billion parameters: Llama-3.1-8B-Instruct \cite{grattafiori2024llama}; its derivatives Vicuna-7B-v1.5 \cite{vicuna2023}, optimized for conversational tasks, and Med-Alpaca-7B \cite{han2023medalpaca}, finetuned for medical question-answering; as well as Phi-3.5-Mini-Instruct \cite{abdin2024phi} and Mistral-7B \cite{jiang2023mistral}. We assess three adaptation techniques: \textbf{1. Prefix Prompting.} The model is prompted using the same instructions employed during training data generation (Appendix~\ref{sec:A_prompt}). \textbf{2. ICL.} The model is given a number of free-form reports along with their structured counterparts. These examples are manually selected from the training set to optimally represent the data distribution. \textbf{3. LoRA Finetuning.} The LLM is finetuned for five epochs on the complete training set using LoRA with a rank of eight, modifying approximately 0.1\% of the model’s parameters by injecting trainable adapters into the key, query, and value projection matrices of the self-attention layers. We use a cosine learning rate scheduler with an initial learning rate of $1e^{-4}$, an effective batch size of 256 and the Adam optimizer. Detailed finetuning configurations are provided in Appendix~\ref{sec:A_peft}.
Throughout the project, we systematically evaluated different combinations of these adaptation techniques. This included varying the number of in-context examples (1-shot, 2-shot) as well as combining \textit{Prefix Prompting} with ICL to assess their complementary effects. We also experimented with hybrid approaches that combined LoRA finetuning with prompting-based methods. However, these configurations did not yield consistent performance gains and introduced substantial overhead in terms of training time and memory usage, primarily due to increased input lengths.

\begin{figure*}[h]
    \centering
    \includegraphics[width=\linewidth]{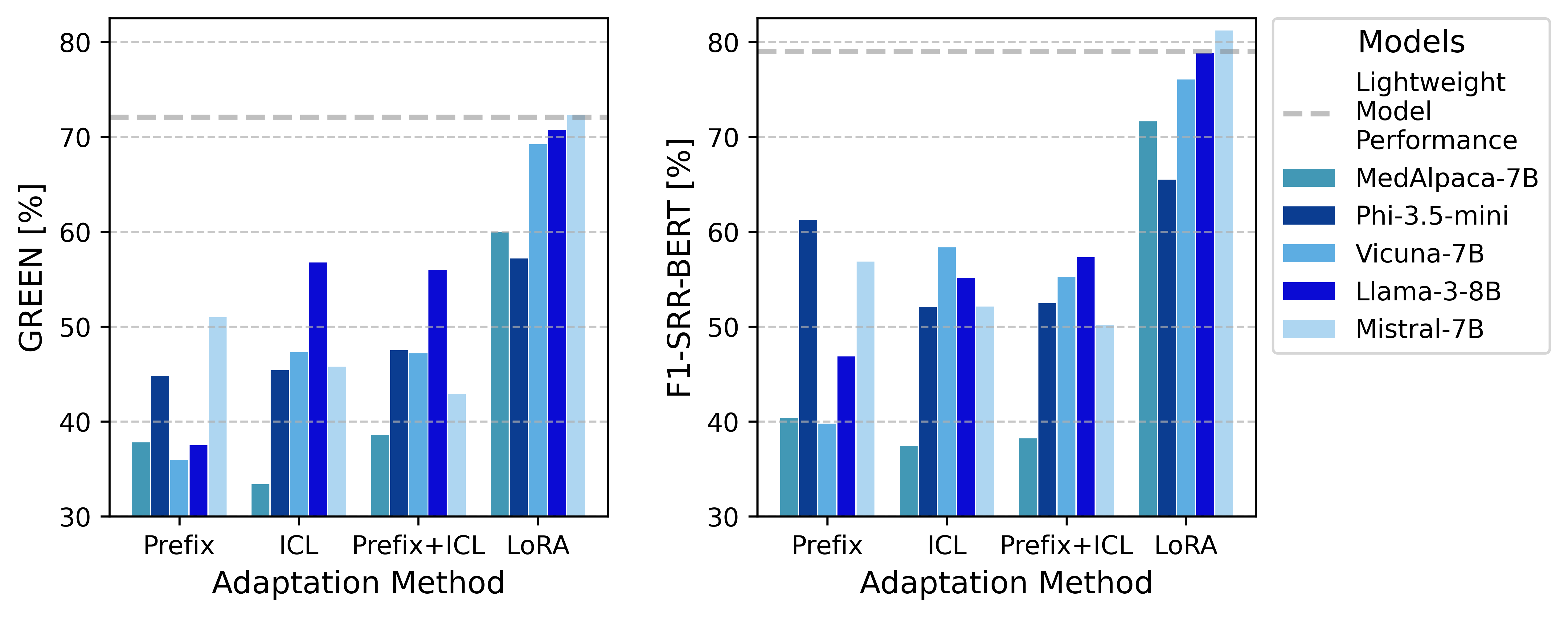}
    \caption{Comparison of LLM Adaptation Methods and the best performing lightweight model (BERT2BERT initialized from RoBERTa-PM-M3). (Left)/(Right) The figure depicts the GREEN Score/F1-SRR-BERT Score for five different LLMs across various adaptation methods, including prefix prompting, in-context learning (ICL), the combination of prefix prompting with ICL, and LoRA finetuning for five epochs.}
    \label{fig:llmadapt}
\end{figure*}

\subsection{Benchmarking Lightweight Models Against LLMs}
\label{sec:bench}
Building on the previous experiment—which compared similarly sized LLMs under various adaptation strategies—we now turn to a scale-sensitive evaluation of our lightweight model. To this end, we benchmark its performance against LLaMA-3 models of increasing size (1B, 3B, 8B, and 70B parameters), leveraging the architectural consistency across this family to isolate the effects of model scale. Each variant is evaluated using the two most effective adaptation strategies identified in our prior experiments: \textit{Prefix+ICL} for prompting-based approaches and \textit{LoRA} for parameter-efficient finetuning. We then compare the computational costs associated with training and deploying the lightweight model, LLaMA-3-3B, and LLaMA-3-70B. This comparison includes the average F1-SRR-BERT score, training time per epoch, inference time per sample, inference costs per sample, and $CO_2$ emissions per sample. Financial costs are estimated using the Google Cloud pricing calculator\footnote{https://cloud.google.com/products/calculator (Assessed January 2025)}, and $CO_2$ emissions are calculated with CodeCarbon \cite{lacoste2019quantifying}. These comparisons provide insights into the trade-offs between large-scale LLMs and compact lightweight models in terms of both performance and resource efficiency.

%% file: 4_Results.tex
The models are evaluated using all metrics introduced in Section~\ref{sec:metrics}. We primarily report results using GREEN and F1-SRR-BERT Score, as they provide the most comprehensive assessments of clinical accuracy and structural consistency. However, unless stated otherwise, the observed trends hold across all metrics. A detailed comparison across all metrics is provided in Appendix~\ref{sec:A_results}. 

\subsection{Comparison of Lightweight Models and Domain Adaptation}
\label{sec:res_expert}
As introduced in Section~\ref{sec:met_lightweight}, we initialized our lightweight models with the weights from different pretrained models. Specifically, we evaluate four different pretrained models as initializations for the BERT2BERT model and five for the T5 model (Tables~\ref{tab:t5_models} and~\ref{tab:roberta_models}). Each pretraining configuration was trained three times with different random seeds. Figure~\ref{fig:t5_domain} presents the model performance for the GREEN and F1-SRR-BERT metrics, while a more comprehensive overview can be found in Table~\ref{tab:metrics_expert}. 
For the BERT2BERT model, domain adaptation shows a clear but non-linear impact on performance. Pretraining on biomedical text improves GREEN by $0.4\%$ over the general-text baseline, while adding radiology reports yields a more substantial $4.5\%$ improvement. However, pretraining exclusively on radiology reports (RadBERT) provides only a marginal $0.3\%$ increase. For the T5 model, instruction-tuning alone leads to $0.3\%$ improvement over the general-text baseline. Pretraining on biomedical text and radiology reports achieves a $2.5\%$ gain, while using exclusively radiology reports leads to $4.4\%$ increase. However, the biomedical text initialization (SciFive) underperforms the general baseline by $2.4\%$. Table~\ref{tab:metrics_expert} confirms that these trends persist across both datasets and sections, with scores for the Impression section being on average by $\approx20\%$ higher. Overall, BERT2BERT models outperform T5 variants, with the best BERT2BERT model (RoBERTa-PM-M3) beating the best T5 (Clin-T5-Base) by $2.6\%$ on GREEN and $1.5\%$ on F1-SRR-BERT. 

\subsection{Adaptation of LLMs}
\label{sec:res_llms}
We present the results of adapting LLMs to the structuring task as outlined in Section~\ref{sec:met_llm}. Figure~\ref{fig:llmadapt} visualizes the average test set performance on the GREEN and F1-SRR-BERT metrics across a selection of the proposed adaptation methods: prefix prompting, 2-shot in-context learning (ICL), the combination of prefix prompting and ICL, and LoRA finetuning. LoRA finetuning consistently achieves the highest performance across all models. The detailed breakdown of results across the structured Findings and Impression sections is provided in Tables~\ref{tab: LLMs_detail} and~\ref{tab: LLMs_comp} of the Appendix.
Averaged across all five LLMs, 2-shot ICL improves performance compared to prefix prompting by $22.2\%/20.6\%$ in GREEN/F1-SRR-BERT on Findings and $9.6\%/-1.0\%$ on Impression. \textit{Prefix+ICL} shows a $77.8\%/79.2\%$ improvement on Findings but also $-5.9\%/-4.1\%$ on Impression.
LoRA finetuning achieves the highest scores overall, outperforming prefix prompting by $263\%/237\%$ on Findings and $8.7\%/6.5\%$ on Impression. 
Across LLMs, Llama-3-8B performs best in ICL methods, while Mistral-7B achieves the highest performance in LoRA finetuning. The overall best-performing configuration is Mistral-7B with LoRA finetuning.

\begin{figure}[!t]
        \centering
        \includegraphics[width=\linewidth]{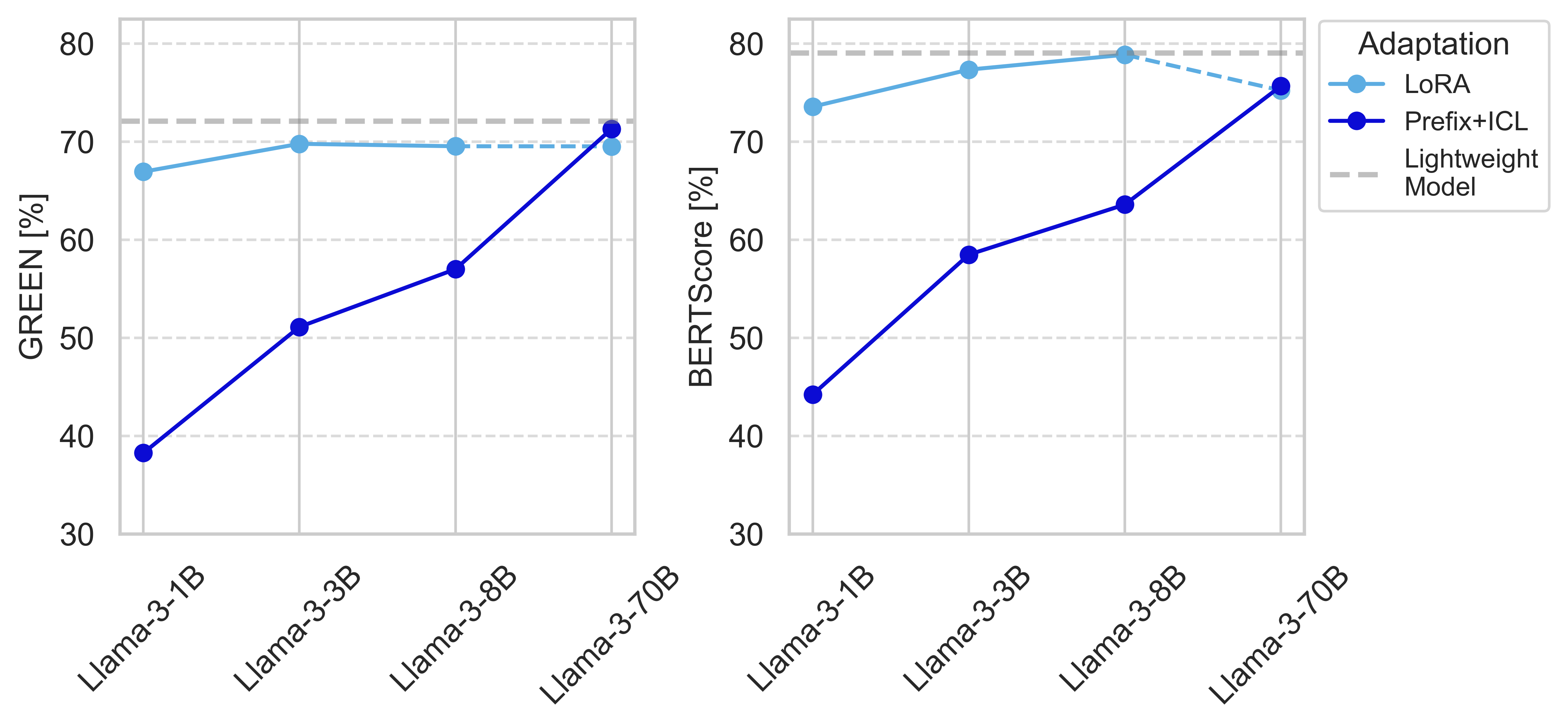}
        \caption{Model performance of LLaMA-3 models of increasing size. (Left/Right) The figure shows the GREEN scores and BERTScores for adaptation using Prefix+ICL and LoRA finetuning, respectively. The result for the LLaMA-3-70B model with LoRA finetuning is indicated with a dashed line, as this configuration was trained for only one epoch—compared to five epochs for the other models—due to computational constraints.}
        \label{fig:benchmark}
\end{figure}

\subsection{Benchmarking}
\label{sec:res_bench}
Building on these results, we benchmark our best lightweight model against LLaMA-3 models of increasing parameter counts. Figure~\ref{fig:benchmark} demonstrates a general positive correlation between the LLM's model size and performance in structuring radiology reports, with the exception of LLaMA-3-70B. Despite being the largest model, it underperforms when adapted via LoRA, likely due to insufficient training. This size-performance trend is more evident with \textit{Prefix+ICL} adaptation. While LLaMA-3-1B achieves only $53.0\%$/$55.9\%$ of the lightweight model's performance (GREEN/F1-SRR-BERT), LLaMA-3-70B reaches $98.9\%$/$95.8\%$. LoRA boosts LLaMA-3-1B to $92.9\%$/$93.0\%$, and enables the larger variants to slightly outperform the lightweight model on the Findings section. However, when averaged across both sections, no LLM surpasses the lightweight model.
Moreover, the relative benefit of LoRA over \textit{Prefix+ICL} diminishes as model size increases, with both methods converging in performance—and LoRA occasionally underperforming—particularly on clinically relevant metrics such as F1-RadGraph, GREEN, and F1-SRR-BERT.
Given these findings, we next turn to a cost analysis. As shown in Table~\ref{tab:costs}, the lightweight model offers considerable advantages in training time, financial cost, and environmental impact—producing only $8.3\%$ and $0.7\%$ of the $CO_2$ emissions of LLaMA-3-3B and 70B, respectively. Inference efficiency follows a similar pattern: even under the least favorable deployment scenario, the lightweight model exhibits up to $91.8\%$ lower latency and $98.4\%$ lower emissions than LLaMA-3-70B. Under optimal conditions, these savings exceed $99.9\%$. 

\subsection{Qualitative Analysis}
\label{sec:res_qual}
To complement the quantitative analysis, Figure~\ref{fig:qualitative} presents a qualitative comparison of BERT2BERT, Mistral-7B, and expert-reviewed reports. Both models successfully adhere to our predefined template (see Figure~\ref{fig:data_training} for reference), particularly in the Findings section, where content is well-aligned with organ system categories. A full test set analysis shows that the lightweight model correctly applies the Findings and Impression section headers in all cases, while the LLM deviates in $5\%$ of instances, occasionally using all capital letters or omitting section names in less than $1\%$ of reports. Both models, as well as expert annotations, generally include only relevant organ systems, but occasionally report less relevant negative findings (e.g., \textit{"Pleura: - No specific findings reported"}). Complete omission of relevant findings occurs in less than $1\%$ of cases, indicating high completeness in capturing clinical details. Differences in prioritization in the Impression section are observed in fewer than $5\%$ of reports for both models, demonstrating occasional variation but overall consistency with expert-reviewed reports.

\begin{table}[h] 
\begin{center}
\caption{Trade-off between model performance and computational costs for training and inference using  total training time [h], C02 emission during training [kg], F1-SRR-BERT Score [\%],  inference time [s/sample], inference cost [\$/sample], and $CO_2$ emissions [mg/sample] across the best-performing BERT2BERT, LLaMA-3-3B, and LLaMA-3-70B models using NVIDIA A100-80GB GPUs.} 
\label{tab:costs}
\scriptsize
\begin{tabular}{c| r| c| c| c}
    \hline
    \multicolumn{2}{r|}{Model} & Lightweight & 3B LLM & 70B LLM$^{\circ}$\\ \hline \hline
    \multicolumn{2}{r|}{\# Parameters} & 0.28B & 3.21B & 70.6B \\ \hline
    \multicolumn{2}{r|}{Training time [h]} & 2.1 & 15.0 & 44.5$^{\circ}$\\ \hline
    \multicolumn{2}{r|}{Training $CO_2$ eq. [kg]} & 0.58 & 7.0 & 82.6$^{\circ}$\\
    \hline 
    \multirow{4}{*}{\rotatebox{90}{Inference}} 
    &SRR-BERT [\%]& 79.1 & 77.4 & 75.2 \\ \cline{2-5}
    & Time [s]& 3.1 (0.16)$^{\ast}$ & 10.7 & 1260 (37.7)$^{\dagger}$ \\ \cline{2-5}
    &Cost [\$] & 0.0043 (2e-4)$^{\ast}$ & 0.015 & 1.76 (0.21)$^{\dagger}$ \\ \cline{2-5}
    &$CO_2$ eq. [g]  & 0.075 (0.0038)$^{\ast}$ & 0.25 & 67.7 (7.9) $^{\dagger}$ \\ \hline
    
    \multicolumn{5}{l}{$^{\circ}$ Only trained for 1 epoch. Trained on four GPUs instead of one.}\\
    \multicolumn{5}{l}{$^{\ast}$ For single-sample (batch-wise) processing.}\\
    \multicolumn{5}{l}{$^{\dagger}$ Executed on 1 (4) NVIDIA A100 (80GB) GPU(s).}\\
\end{tabular}
\end{center}
\end{table}

%% file: 5_Discussion.tex
In this paper, we propose lightweight, task-specific models for structuring radiology reports into a predefined template. Despite being 10–250 times smaller than finetuned LLMs, our models achieved comparable performance while offering significant advantages in speed, cost-efficiency, and sustainability. To enable large-scale supervised training, we leveraged GPT-4 as a weak annotator to generate a training dataset, aligning chest radiology reports from MIMIC-CXR and CheXpert Plus with their corresponding structured versions as ground truth. Since GPT-generated data may contain inconsistencies and biases, we evaluated all models on a human-reviewed test set. Our study focused on two types of lightweight models, BERT2BERT and T5. Overall, our BERT2BERT model performed best when initialized from RoBERTa-PM-M3, surpassing the best T5 variant, Clin-T5-Base, by $2.6\%$ on GREEN. Our results further indicate that pretraining on biomedical texts - particularly radiology reports - generally improved model performance. However, despite being pretrained exclusively on radiology reports, the RadBERT model did not outperform general-text variants. This suggests that pretraining factors beyond the training corpus, such as architectural choices and optimization techniques, may also influence model performance. For example, RoBERTa-PM-M3 benefited from a distillation process from RoBERTa-large-PM-M3-Voc.

To balance performance with computational feasibility, we first restricted our comparison to LLMs within the 3-8B parameter tier, evaluating different adaptation techniques within this range. We showed that LoRA finetuning consistently outperformed prefix prompting and ICL methods. As shown in Table~\ref{tab: LLMs_comp}, this trend was primarily driven by performance differences on the Findings section. Given that our evaluation assessed each organ system independently and assigned zero points to missing or inconsistently labeled headers (e.g., \textit{'Lungs and Airways'} vs. \textit{'Lungs'}), the results suggest that LoRA finetuning more effectively aligned LLM outputs with the predefined reporting template. We believe that although organ system names are provided in both the prefix prompt (see Appendix~\ref{sec:A_prompt}) and the ICL examples, the absence of iterative feedback mechanisms in these methods made it challenging for models to internalize and consistently enforce correct structured formatting. 

Among the five evaluated LLMs and four adaptation techniques, Mistral-7B and LLaMA-3-8B achieved the best results. Notably, MedAlpaca-7B underperformed compared to general-domain models of similar size, suggesting that current medicine-specific LLMs may not yet offer clear advantages for structured report generation. We selected LLaMA-3 models with 1B, 3B, 8B, and 70B parameters for benchmarking our lightweight model against LLMs of increasing size in Section~\ref{sec:res_bench}. Under the two most effective adaptation strategies—\textit{Prefix+ICL} and LoRA—performance generally improved with model size, with LoRA finetuning ultimately enabling larger models to surpass the lightweight model on the Findings section. This came, however, at the cost of significantly longer training times and higher inference costs. 

Our qualitative analysis in Section~\ref{sec:res_qual} showed that both models (the lighweight model and Mistral-7B LLM finetuned with LoRA) followed the predefined template when tested on expert-annotated reports, omitting relevant findings in less than $1\%$ of cases. This suggests that lightweight models (<300M parameters) can effectively learn structured formatting while maintaining clinical accuracy. Furthermore, the results indicate that our GPT-generated annotations provided a sufficient training signal, though expert review remains crucial for ensuring data reliability.

%% file: 6_Conclusion.tex
We demonstrate that lightweight, task-specific models with less than 300M parameters can effectively structure radiology reports according to a predefined template, providing a practical and scalable alternative to LLMs, while addressing concerns around computational efficiency, data privacy, and deployment feasibility. Our best-performing lightweight model, a BERT2BERT architecture initialized from two pretrained RoBERTa-PM-M3 models, achieved competitive performance while maintaining a significantly lower computational footprint. While LLaMA-3 variants with more than 3 billion parameters achieved slightly better performance on the Findings section when finetuned with LoRA, the lightweight model operated at less than $25\%$ of their inference cost and $CO_2$ emissions, making it a more resource-efficient solution. These findings reinforce the lightweight model’s viability for real-world clinical applications, where infrastructure limitations, privacy regulations, and sustainability concerns play a critical role.

%% file: Appendix.tex
\subsection{GPT-4 prompt template for structuring of radiology reports}
\label{sec:A_prompt}
The following prompt was executed with GPT-4 "Turbo 1106 preview" via Azure services to structure free-text radiology reports according to our template. The account was explicitly opted out of human review.
\begin{lstlisting}[style=promptstyle]
Your task is to improve the formatting of a radiology report to a clear and 
concise radiology report with section headings. 
Guidelines: 
    1. Section Headers: Each section should start with the section header 
    followed by a colon. Provide the relevant information as specified for 
    each section. 
    2. Identifiers: Remove sentences where identifiers have been replaced 
    with consecutive underscores ('\_\_\_'). 
    3. Findings and Impression Sections: Focus solely on the current 
    examination results. Do not reference previous studies or historical data. 
    4. Content Restrictions: Strictly include only the content that is relevant 
    to the structured sections provided. Do not add or extrapolate information 
    beyond what is found in the original report. If the original report doesn't 
    contain the information necessary to generate a section, write the section 
    header and then leave the section empty. Do not make up any findings.! 
    Sections to include (if applicable):
    1. Exam Type: Provide the specific type of examination conducted. 
    2. History: Provide a brief clinical history and state the clinical 
    question or suspicion that prompted the imaging. 
    3. Technique: Describe the examination technique and any specific protocols 
    used. 
    4. Comparison: Note any prior imaging studies reviewed for comparison with 
    the current exam. 
    5. Findings: 
        Describe all positive observations and any relevant negative 
        observations for each organ or organ system under distinct headers.
        Start with the organ system name followed by a colon, then list 
        observations. 
        Here is the corresponding template: 
            Organ 1: 
                - Observation 1 
            Organ 2:
                - Observation 1 
                - Observation 2 
    Use only the following headers for organ systems:
    - Lungs and Airways
    - Pleura
    - Cardiovascular 
    - Hila and Mediastinum 
    - Tubes, Catheters, and Support Devices 
    - Musculoskeletal and Chest Wall
    - Abdominal
    - Other
    6. Impression: Summarize the key findings with a numbered list from 
    the most to the least clinically relevant. Ensure all findings are numbered.
The radiology report to improve is the following: \{report\}
\end{lstlisting}

\subsection{Overview of model checkpoints and pre-training data}
\label{sec:A_checkpoints}

\begin{table}[H]
    \caption{Pretrained T5 models used for initialization along with details of their pretraining corpus.}
    \label{tab:t5_models}
    \centering
    \begin{tabular}{|p{6.2cm}|p{8.8cm}|}
    \hline
    \textbf{Model}  & \textbf{Description}\\ \hline
    \textbf{T5-BASE}~\cite{raffel2020exploring} & Original model, pre-trained on C4. \\ \hline
    \textbf{FLAN-T5-BASE}~\cite{chung2024scaling} & Additional instruction-prompt tuning.\\ \hline
    \textbf{SCIFIVE}~\cite{phan2021scifive} & Fine-tuned on PubMed Abstract~\cite{ncbi_pubmed_1996}, \\
     & and PubMed Central~\cite{ncbi_pubmed_2000}. \\ \hline
    \textbf{CLIN-T5-SCI} & Fine-tuned on PubMed, MIMIC-III~\cite{johnson2016mimic}, \\ \cite{lehman2023clinical}& and MIMIC-IV~\cite{johnson2020mimic}. \\ \hline
    \textbf{CLIN-T5-BASE}~\cite{lehman2023clinical} & Fine-tuned on MIMIC-III and MIMIC-IV. \\ \hline
    \end{tabular}
    
\end{table}

\begin{table}[H]
    \caption{Pretrained RoBERTa models used for initialization of the BERT2BERT model along with details of their pretraining corpus.}
    \label{tab:roberta_models}
    \centering
    \begin{tabular}{|p{6.2cm}|p{8.8cm}|}
    \hline
    \textbf{Model} & \textbf{Description}  \\ \hline
    \textbf{RoBERTa-base}~\cite{liu2019roberta}  & Baseline version, pretrained on Books and Wikipedia.  \\ \hline
    \textbf{BioMed-RoBERTa}~\cite{gururangan2020don} & Pretrained on PubMed abstracts and PubMed Central.\\ \hline
    \textbf{RoBERTa-base-PM-M3-Voc-distill-} & Pretrained on PubMed abstracts, PubMed Central \\
    \textbf{align} \cite{lewis2020pretrained} & full-text articles, and MIMIC-III.         \\ \hline
    \textbf{RadBERT-RoBERTa}~\cite{yan2022radbert}   & Fine-tuned on radiology reports from the Veterans \\ & Affairs health care system. \\ \hline
    \end{tabular}
    
\end{table}

\twocolumn
\subsection{Considerations and hyperparameters for end-to-end training}
\label{sec:A_endtoend}
We train all expert models (BERT2BERT and T5 instances) with the following set of hyperparameters:
\begin{itemize}
    \item Cosine learning rate scheduler, starting at $1e^{-4}$, with $5\%$ warm-up ratio before decay.
    \item Maximum of 10 epochs, with early stopping enabled by loading the best model at the end based on validation performance.
    \item Batch size of 32 per device for training and 16 for evaluation, with four gradient accumulation steps, resulting in an effective batch size of 128 for training.
    \item Adam optimizer with $\beta_2 = 0.95$ and weight decay of 0.1.
    \item Sequence lengths: Model processes a maximum input length of 370 tokens, with generated outputs constrained between 120 and 286 tokens.
\end{itemize}
We experimented with different learning rate schedulers and initial learning rates but found the here presented set to give better performance in the validation loss.

\subsection{Considerations and hyperparameters for parameter-efficient fine-tuning}
\label{sec:A_peft}
As discussed in Section~\ref{sec:met_llm}, we initially finetune all LLMs using the same hyperparameters. We apply LoRA and adjust the target modules to align with each LLM’s architecture. We find that, due to their comparable size, using the same LoRA rank and scaling factor leads to a similar proportion of updated parameters across all models ($ \sim 0.1\%$). We use the following set of hyperparameters:
\begin{itemize}
    \item Cosine learning rate scheduler, starting at $1e^{-4}$, with $5\%$ warm-up ratio before decay.
    \item Maximum of 5 epochs, with early stopping enabled by loading the best model at the end based on validation performance.
    \item LoRA adaptation with rank $r=8$ and scaling factor $\alpha = 8$ to enable parameter-efficient fine-tuning.
    \item Batch size of 16 per device for training and 1 for evaluation, with 16 gradient accumulation steps, resulting in an effective training batch size of 256.
    \item Adam optimizer with $\beta_2 = 0.95$ and weight decay of 0.1.
\end{itemize}
We use similar settings as in expert model fine-tuning but reduce the maximum number of epochs due to computational constraints. The results in Section~\ref{sec:res_bench} later confirm our initial estimate for the optimal LoRA rank.

\newpage
\onecolumn
\subsection{Detailed Evaluations of Model Performance}
\label{sec:A_results}
\begin{table}[h]
    \centering
    \small
    \caption{Detailed comparison of expert models. This table presents test set evaluations of our finetuned expert models initialized from different pre-trained checkpoints. Each model was trained three times with different random seeds and evaluated on the Findings sections of the MIMIC ($F_M$) and CheXpert ($F_C$) test sets, as well as their corresponding Impression sections ($I_M$ and $I_C$).\\
    }
    \begin{tabular}{lccccccc}
        \hline \rowcolor[HTML]{EFEFEF}
        \textbf{Model} & \textbf{Section} & \textbf{BLEU} & \textbf{ROUGE-L} & \textbf{BERTScore} & \textbf{RadGraph} & \textbf{GREEN} & \textbf{SRR-BERT} \\
        
        \hline
        BERT2BERT &&&&&&& \\
        \hline
        roberta-base& $F_M$  & 31.3	&62.2&	\cellcolor[HTML]{C0C0C0}\textbf{67.4}&	54.8	&66.1&	\cellcolor[HTML]{C0C0C0}\textbf{73.0}\\
                    & $F_C$  & 30.6&	59.0&	64.7&	50.1	&63.0&	69.4\\
                    & $I_M$  & 41.1&	65.4&	79.7&	57.5	&65.6&	\cellcolor[HTML]{C0C0C0}\textbf{81.8}\\
                    & $I_C$  & 51.1&	74.9&	86.3&	66.1&82.0&		94.5\\
        \hline
        roberta-biomed& $F_M$  & 31.6&	60.4&	65.4&	53.1	&62.8&	70.4\\
                    & $F_C$  & 29.4	&57.8&	63.8&	48.2	&62.1&	70.0\\
                    & $I_M$  & 34.0&	65.5&	79.9&	58.0	&69.1&	\cellcolor[HTML]{C0C0C0}\textbf{81.8}\\
                    & $I_C$  & 48.3&	74.1&	86.1&	65.3&82.0&		91.3\\
        \hline
        roberta-PM& $F_M$  & \cellcolor[HTML]{C0C0C0}\textbf{33.3}& \cellcolor[HTML]{C0C0C0}\textbf{62.6}& \cellcolor[HTML]{C0C0C0}\textbf{67.4}& 54.3& \cellcolor[HTML]{C0C0C0}\textbf{67.0}& 71.9 \\
                    & $F_C$  & \cellcolor[HTML]{C0C0C0}\textbf{32.8}& \cellcolor[HTML]{C0C0C0}\textbf{62.5}&\cellcolor[HTML]{C0C0C0} \textbf{67.3}& \cellcolor[HTML]{C0C0C0}\textbf{53.8}&\cellcolor[HTML]{C0C0C0}\textbf{64.2} & \cellcolor[HTML]{C0C0C0}\textbf{72.8}\\
                    & $I_M$  & 42.0& 66.1& 79.8& 56.5& \cellcolor[HTML]{C0C0C0}\textbf{71.8}& 81.4\\
                    & $I_C$  & \cellcolor[HTML]{C0C0C0}\textbf{53.4}& \cellcolor[HTML]{C0C0C0}\textbf{77.6}& \cellcolor[HTML]{C0C0C0}\textbf{87.5}& 67.7& 86.4& 90.1\\
        \hline
        roberta-rad& $F_M$  & 32.6&	62.1&	66.8&	\cellcolor[HTML]{C0C0C0}\textbf{54.9}	&64.8&	71.8\\
                    & $F_C$  & 29.4&	59.2	&64.2&	50.7&61.0&	69.1\\
                    & $I_M$  & \cellcolor[HTML]{C0C0C0}\textbf{42.3}	&\cellcolor[HTML]{C0C0C0}\textbf{67.5}&	\cellcolor[HTML]{C0C0C0}\textbf{80.6}&	\cellcolor[HTML]{C0C0C0}\textbf{58.9}&69.7&	81.7\\
                    & $I_C$  & 52.4	&76.6&	87.2&	65.7&86.7&	94.3\\
        \hline
        T5 &&&&&&& \\
        \hline
        T5-Base     & $F_M$  & 26.4 & 52.8 & 58.8 & 64.9 & 58.6 & 63.6	\\
                    & $F_C$  & 	26.0 & 57.2 & 61.9 & 49.1 & 59.7 & 66.5	\\
                    & $I_M$  & 	35.8 & 61.7 & 77.7 & 56.2 & 69.8 & 80.1\\
                    & $I_C$  & 48.5 & 73.2 & 85.8 & \cellcolor[HTML]{C0C0C0}\textbf{67.9} & 81.2 & 87.1	\\
        \hline
        Flan-T5-Base & $F_M$  & 27.9 & 55.9 & 61.0 & 48.0 & 59.3 & 65.4	\\
                    & $F_C$  & 	30.3 & 59.2 & 63.5 & 51.1 & 62.2 & 66.2 \\
                    & $I_M$  & 37.3 & 62.0 & 77.6 & 55.5 & 66.2 & 77.8\\
                    & $I_C$  & 51.6 & 76.1 & 87.1 & 68.6 & 82.3 & 91.7\\
        \hline
        SciFive     & $F_M$  & 24.1 & 49.3 & 55.6 & 43.4 & 56.4 & 62.0\\
                    & $F_C$ &  24.6 & 54.1 & 60.5 & 47.2 & 56.7 & 65.7	\\
                    & $I_M$  & 38.6 & 63.2 & 78.8 & 59.5 & \cellcolor[HTML]{C0C0C0}\textbf{71.8} & 82.9	\\
                    & $I_C$  & 46.8 & 71.4 & 85.1 & 68.1 & 77.8 & 89.4	\\
        \hline
        Clin-T5-Sci & $F_M$ &  28.7 & 59.0 & 64.4 & 50.7 & 62.4 & 68.9	\\
                    & $F_C$  & 23.4 & 52.5 & 57.1 & 44.0 & 56.1 & 62.0	\\
                    & $I_M$  & 33.6 & 59.4 & 76.2 & 51.4 & 63.8 & 76.3\\
                    & $I_C$  & 46.7 & 71.8 & 84.6 & 62.8 & 84.0& 93.0	\\
        \hline
        Clin-T5-Base & $F_M$  & 29.8 & 58.3 & 64.0 & 50.9 & 62.7 & 68.6\\
                    & $F_C$ & 	27.1 & 57.3 & 62.0 & 49.0 & 60.9 & 68.1	\\
                    & $I_M$ & 	37.6 & 63.3 & 78.9 & 55.7 & 68.7 & 80.2 \\
                    & $I_C$  &  48.4 & 74.8 & 85.5 & \cellcolor[HTML]{C0C0C0}\textbf{67.9} & \cellcolor[HTML]{C0C0C0}\textbf{88.8} & \cellcolor[HTML]{C0C0C0}\textbf{94.6}	\\
        
        \hline
    \end{tabular}
    \label{tab:metrics_expert}
\end{table}

\begin{table}[]
    \centering
    \small
    \caption{Comparison of LLM performance across different adaptation and finetuning methods. Results are averaged over all samples in the expert-reviewed MIMIC and CheXpert test sets and reported separately for the Findings and Impression sections. The highest score for each model across adaptation techniques is highlighted.}
    \begin{tabular}{llcccccc}
        \hline \rowcolor[HTML]{EFEFEF}
        \textbf{Model} & \textbf{Method} & \textbf{BLEU} & \textbf{ROUGE-L} & \textbf{BERTScore} & \textbf{Radgraph} & \textbf{GREEN} & \textbf{F1-Score} \\
        \hline
        \multicolumn{8}{c}{\textbf{Findings Section}} \\
        \hline
        Medalpaca-7B& Prefix& 0.0 & 0.0 & 0.0 & 0.0 & 0.0& 0.0\\ 
                    & 1-shot ICL&0.0 & 0.2 & 1.4 & 0.1 & 0.1  & 0.9 \\
                    & 2-shot ICL ICL& 0.0 & 0.0 & 0.0 & 0.0 & 0.0& 0.0\\
                    & Prefix+ICL & 0.0 & 2.3 & 7.6 & 0.7 & 11.4 & 5.4\\
                    & LoRA & \cellcolor[HTML]{C0C0C0}\textbf{19.7} & \cellcolor[HTML]{C0C0C0}\textbf{45.4} & \cellcolor[HTML]{C0C0C0}\textbf{50.5} & \cellcolor[HTML]{C0C0C0}\textbf{41.3} & \cellcolor[HTML]{C0C0C0}\textbf{51.0} & \cellcolor[HTML]{C0C0C0}\textbf{57.1}\\
        \hline
        Phi-3.5-mini     & Prefix&11.0 & 34.6 & 38.9 & 26.7 & 38.1 & 46.5\\
                    & 1-shot ICL&8.6 & 21.5 & 24.8 & 20.1 & 25.6 & 26.4 \\
                    & 2-shot ICL&6.8 & 20.1 & 24.1 & 18.5 & 23.2 & 25.8\\
                    & Prefix+ICL & 14.3&35.3&40.7&28.8&38.3&43.6\\
                    & LoRA &\cellcolor[HTML]{C0C0C0}\textbf{17.8} & \cellcolor[HTML]{C0C0C0}\textbf{43.8} & \cellcolor[HTML]{C0C0C0}\textbf{49.5} & \cellcolor[HTML]{C0C0C0}\textbf{39.0} & \cellcolor[HTML]{C0C0C0}\textbf{46.7} & \cellcolor[HTML]{C0C0C0}\textbf{52.9}\\
        \hline
        Vicuna-7B   & Prefix& 0.0 & 0.0 & 0.0 & 0.0 & 0.0& 0.0\\ 
                    & 1-shot ICL& 5.9 & 21.5 & 29.2 & 17.5 & 22.8 & 32.4 \\
                    & 2-shot ICL& 7.1 & 19.8 & 24.6 & 17.0 & 22.6 & 28.2 \\
                    & Prefix+ICL&7.4 & 23.7 & 30.9&19.0& 26.3 & 32.2\\
                    & LoRA & \cellcolor[HTML]{C0C0C0}\textbf{32.7} & \cellcolor[HTML]{C0C0C0}\textbf{62.1} & \cellcolor[HTML]{C0C0C0}\textbf{66.8} & \cellcolor[HTML]{C0C0C0}\textbf{54.2} & \cellcolor[HTML]{C0C0C0}\textbf{66.1} & \cellcolor[HTML]{C0C0C0}\textbf{70.6} \\

        \hline       
        LLaMA-3-8B    & Prefix& 2.4 & 10.9 & 12.8 & 8.6 & 13.1 & 12.7 \\
                    & 1-shot ICL& 13.1 & 35.6 & 42.1 & 30.6 & 40.1 & 46.4 \\
                    & 2-shot ICL& 13.7 & 36.4 & 42.1& 31.1 & 38.0 & 46.4 \\
                    & Prefix+ICL&18.7&44.7&51.1&37.6&48.6 & 56.6\\
                    & LoRA & \cellcolor[HTML]{C0C0C0}\textbf{35.0} & \cellcolor[HTML]{C0C0C0}\textbf{62.9} & \cellcolor[HTML]{C0C0C0}\textbf{68.4} & \cellcolor[HTML]{C0C0C0}\textbf{54.4} & \cellcolor[HTML]{C0C0C0}\textbf{68.1} & \cellcolor[HTML]{C0C0C0}\textbf{74.0} \\
        \hline
        Mistral-7B &  Prefix& 8.2 & 26.8 & 30.3 & 6.9 & 32.5 & 35.8 \\
                    & 1-shot ICL& 6.5 & 15.2 & 18.4 & 14.7 & 16.9 & 19.4\\
                    & 2-shot ICL& 5.9 & 14.9 & 18.1 & 12.5 & 18.5 & 18.4 \\
                    & Prefix+ICL&14.3&30.6&35.6&24.8& 34.1& 38.9\\
                    & LoRA & \cellcolor[HTML]{C0C0C0}\textbf{37.5} & \cellcolor[HTML]{C0C0C0}\textbf{69.3} & \cellcolor[HTML]{C0C0C0}\textbf{73.6} & \cellcolor[HTML]{C0C0C0}\textbf{61.2} & \cellcolor[HTML]{C0C0C0}\textbf{72.4} & \cellcolor[HTML]{C0C0C0}\textbf{77.7}\\
        \hline
        \multicolumn{8}{c}{\textbf{Impression Section}} \\
        \hline
        Medalpaca-7B& Prefix& 23.6 & 55.1 & 63.9 & 52.0 & 75.6 & 80.8\\
                    & 1-shot ICL& 23.3 & 54.0 & 60.7 & 50.3 & 66.8 & 74.1\\
                    & 2-shot ICL& \cellcolor[HTML]{C0C0C0}\textbf{25.8} & \cellcolor[HTML]{C0C0C0}\textbf{56.5} & \cellcolor[HTML]{C0C0C0}\textbf{66.7} & \cellcolor[HTML]{C0C0C0}\textbf{57.4} & \cellcolor[HTML]{C0C0C0}\textbf{77.2} & \cellcolor[HTML]{C0C0C0}\textbf{76.5}\\
                    & Prefix+ICL & 18.4 & 46.7 & 60.8 & 39.8 & 65.2 & 63.8 \\
                    & LoRA & 17.4 & 53.5 & 63.4 & 38.4 & 68.9 & 86.2\\
        \hline
        Phi-3.5-mini     & Prefix& 19.2 & 45.7 & 63.7 & 43.7 & 51.5 & 76.0 \\
                    & 1-shot ICL& 24.4 & 48.6 & 66.8 & 47.7 & 65.3 & 77.8\\
                    & 2-shot ICL& 32.6 & 48.5 & 66.8 & 51.9 & \cellcolor[HTML]{C0C0C0}\textbf{71.8} & \cellcolor[HTML]{C0C0C0}\textbf{79.2}\\
                    & Prefix+ICL&27.1&52.5&69.7&46.7&64.2&74.2\\
                    & LoRA &\cellcolor[HTML]{C0C0C0}\textbf{39.3}	&\cellcolor[HTML]{C0C0C0}\textbf{64.4}	&\cellcolor[HTML]{C0C0C0}\textbf{77.3}	&\cellcolor[HTML]{C0C0C0}\textbf{56.2}	&67.5&	78.1\\
        \hline
        Vicuna-7B   & Prefix& 34.0 &64.8 & 73.7 & 57.8 & 71.9 & 79.6\\
                    & 1-shot ICL& \cellcolor[HTML]{C0C0C0}\textbf{38.8} & 64.7 & \cellcolor[HTML]{C0C0C0}\textbf{77.5} & \cellcolor[HTML]{C0C0C0}\textbf{61.5} & 71.9 & \cellcolor[HTML]{C0C0C0}\textbf{84.3}\\
                    & 2-shot ICL& 36.8 & 62.9 & 76.8 & 59.5 & 71.8 & 82.3\\
                    & Prefix+ICL&37.7&\cellcolor[HTML]{C0C0C0}\textbf{64.9}&77.0&56.6& 70.1 &81.4\\
                    & LoRA &38.0 & 63.7 & 70.9 & 54.3 & \cellcolor[HTML]{C0C0C0}\textbf{72.4} & 81.5\\
        \hline
        LLaMA-3-8B    & Prefix& 25.5 & 55.4 & 70.7 & 51.3 & 61.9 & 77.5\\
                    & 1-shot ICL& 9.7 & 27.5 & 45.6 & 33.1 & 73.5 & 63.9\\
                    & 2-shot ICL& 10.6 & 30.1 & 49.3 & 32.6 & 74.0 & 68.2\\
                    & Prefix+ICL& 15.9&45.3&62.5&41.9& 65.4& 70.6 \\
                    & LoRA & \cellcolor[HTML]{C0C0C0}\textbf{35.3} & \cellcolor[HTML]{C0C0C0}\textbf{65.3} & \cellcolor[HTML]{C0C0C0}\textbf{72.0} & \cellcolor[HTML]{C0C0C0}\textbf{54.7} & \cellcolor[HTML]{C0C0C0}\textbf{74.2} & \cellcolor[HTML]{C0C0C0}\textbf{83.7}\\
        \hline
        Mistral-7B  & Prefix& 33.6 & 63.4 & 78.4 & 56.0 & 69.5 & 78.0\\
                    & 1-shot ICL& 38.3 & 65.6 & 76.2 & \cellcolor[HTML]{C0C0C0}\textbf{62.9} & 67.4 & 82.0\\
                    & 2-shot ICL& 39.2 & 66.0 & 77.2 & \cellcolor[HTML]{C0C0C0}\textbf{62.9} & 67.4 & 82.0\\
                    & Prefix+ICL&\cellcolor[HTML]{C0C0C0}\textbf{42.6}&\cellcolor[HTML]{C0C0C0}\textbf{70.7}&\cellcolor[HTML]{C0C0C0}\textbf{80.2}&61.9& 55.0&\cellcolor[HTML]{C0C0C0}\textbf{86.1}\\
                    & LoRA & 42.3 & 67.6 & 74.8 & 57.0 & \cellcolor[HTML]{C0C0C0}\textbf{76.1} & 84.8\\
        \hline
    \end{tabular}
    \label{tab: LLMs_detail}
\end{table}

\newpage

\begin{table}[]
    \centering
    \small
    \caption{Detailed comparison of LLM adaptation methods for the Findings and Impression sections. The table shows average values across all five LLMs (excluding GPT-4), along with percentage changes relative to performance under prefix prompting.}
    \begin{tabular}{lcccccc}
        \hline \rowcolor[HTML]{EFEFEF}
        \textbf{Method} & \textbf{BLEU} & \textbf{ROUGE-L} & \textbf{BERTScore} & \textbf{Radgraph} & \textbf{GREEN} & \textbf{F1-SRR-BERT} \\
        \hline
        \multicolumn{7}{c}{\textbf{Findings Section}} \\
        \hline
        Prefix & 4.31	&14.4	&16.4	&8.43	&16.7	&19.7 \\ 
        1-shot ICL & 6.79	&18.8	&23.2	&16.6	&21.1	&25.1 \\
        &       \textcolor{ForestGreen}{$\uparrow$57.5\%} & \textcolor{ForestGreen}{$\uparrow$30.2\%} & \textcolor{ForestGreen}{$\uparrow$41.4\%} & \textcolor{ForestGreen}{$\uparrow$96.8\%} & \textcolor{ForestGreen}{$\uparrow$26.1\%} & \textcolor{ForestGreen}{$\uparrow$27.3\%} \\ 
        2-shot ICL & 6.67	&18.2	&21.8	&15.8	&20.4 & 23.8 \\ 
        &       \textcolor{ForestGreen}{$\uparrow$54.8\%}  & \textcolor{ForestGreen}{$\uparrow$26.3\%} & \textcolor{ForestGreen}{$\uparrow$33.0\%} &\textcolor{ForestGreen}{$\uparrow$87.4\%}&\textcolor{ForestGreen}{$\uparrow$22.2\%} &\textcolor{ForestGreen}{$\uparrow$20.6\%} \\ 
        Prefix+ICL &11.0&27.3&33.2&22.2&29.7&35.3 \\
        &\textcolor{ForestGreen}{$\uparrow$155\%} &\textcolor{ForestGreen}{$\uparrow$89.6\%} &\textcolor{ForestGreen}{$\uparrow$102\%} &\textcolor{ForestGreen}{$\uparrow$163\%} &\textcolor{ForestGreen}{$\uparrow$77.8\%}&\textcolor{ForestGreen}{$\uparrow$79.2\%}  \\
        LoRA & 28.5	&56.7	&61.7	&50.0	&60.7	&66.5 \\
        &       \textcolor{ForestGreen}{$\uparrow$562\%} & \textcolor{ForestGreen}{$\uparrow$293\%} & \textcolor{ForestGreen}{$\uparrow$277\%} & \textcolor{ForestGreen}{$\uparrow$493\%} & \textcolor{ForestGreen}{$\uparrow$263\%} & \textcolor{ForestGreen}{$\uparrow$237\%} \\ 
        \hline
        \multicolumn{7}{c}{\textbf{Impression Section}} \\
        \hline
        Prefix & 27.2	&56.9	&70.1	&52.1	&66.1	&78.4 \\ 
        1-shot ICL & 26.9	&52.0	&65.3	&50.7	&70.4	&77.0 \\
        &       \textcolor{red}{$\downarrow$-1.1\%} & \textcolor{red}{$\downarrow$-8.5\%} & \textcolor{red}{$\downarrow$-6.8\%} & \textcolor{red}{$\downarrow$-2.7\%} & \textcolor{ForestGreen}{$\uparrow$6.5\%} & \textcolor{red}{$\downarrow$-11.8\%} \\ 
        2-shot ICL & 26.8	&52.8	&67.3	&52.8	&72.4	&77.6 \\
        &       \textcolor{red}{$\downarrow$-1.5\%} & \textcolor{red}{$\downarrow$-7.2\%} & \textcolor{red}{$\downarrow$-3.9\%} & \textcolor{ForestGreen}{$\uparrow$1.3\%} & \textcolor{ForestGreen}{$\uparrow$9.6\%} & \textcolor{red}{$\downarrow$-1.0\%} \\ 
        Prefix+ICL &28.4&56.0&70.0&49.4&62.2&75.2 \\
        &\textcolor{ForestGreen}{$\uparrow$4.4\%}&\textcolor{red}{$\downarrow$-1.6\%}&+0.0\%&\textcolor{red}{$\downarrow$-5.2\%}&\textcolor{red}{$\downarrow$-5.9\%}&\textcolor{red}{$\downarrow$-4.1\%} \\
        LoRA & 34.4	&62.9	&71.6	&52.1	&71.8	&83.5 \\
        &       \textcolor{ForestGreen}{$\uparrow$26.8\%} & \textcolor{ForestGreen}{$\uparrow$10.6\%} & \textcolor{ForestGreen}{$\uparrow$2.2\%} & +0.0\% & \textcolor{ForestGreen}{$\uparrow$8.7\%} & \textcolor{ForestGreen}{$\uparrow$6.5\%} \\
        \hline
    \end{tabular}
    \label{tab: LLMs_comp}
\end{table}
    
\newpage
\begin{table}[]
    \centering
    \small
    \caption{Comparison of lightweight and LLM model performance. Results are averaged over all samples in the expert-reviewed MIMIC and CheXpert test sets and reported separately for the Findings and Impression sections. The highest score for each model across adaptation techniques is highlighted.}
    \begin{tabular}{llcccccc}
        \hline \rowcolor[HTML]{EFEFEF}
        \textbf{Model} & \textbf{Method} & \textbf{BLEU} & \textbf{ROUGE-L} & \textbf{BERTScore} & \textbf{Radgraph} & \textbf{GREEN} & \textbf{F1-Score} \\
        \hline
        \multicolumn{8}{c}{\textbf{Findings Section}} \\
        \hline
        BERT2BERT& Full Training & 32.9 & 62.6 & 67.4 & 54.0 & 66.4 & 72.3\\
        \hline
        LLaMA-3-1B  & Prefix+ICL& 3.7&11.6&17.3&12.2&11.9&16.5\\
                    & LoRA & 29.8&58.8&64.0&50.5&62.3&67.9\\
        \hline
        LLaMA-3-3B  & Prefix+ICL& 10.9&29.6&36.4&24.7&33.3&40.8\\
                    & LoRA & 33.4&\cellcolor[HTML]{C0C0C0}\textbf{65.6}&\cellcolor[HTML]{C0C0C0}\textbf{69.8}&\cellcolor[HTML]{C0C0C0}\textbf{54.6}&\cellcolor[HTML]{C0C0C0}\textbf{68.8}&\cellcolor[HTML]{C0C0C0}\textbf{75.4}\\
        \hline       
        LLaMA-3-8B  & Prefix+ICL&18.7&44.7&51.1&37.6& 48.6& 56.6\\
                    & LoRA & \cellcolor[HTML]{C0C0C0}\textbf{35.0} & 62.9 &68.4 & 54.4 & 68.1& 74.0 \\
        \hline
        LLaMA-3-70B& Prefix+ICL&25.4&53.3&60.2&41.3&53.4&63.1\\
                    & LoRA &30.2&59.1&64.2&51.2&63.3&68.9\\
        \hline
        \multicolumn{8}{c}{\textbf{Impression Section}} \\
        \hline
        BERT2BERT& Full Training & \cellcolor[HTML]{C0C0C0}\textbf{47.7} & \cellcolor[HTML]{C0C0C0}\textbf{71.9} & \cellcolor[HTML]{C0C0C0}\textbf{83.7} & \cellcolor[HTML]{C0C0C0}\textbf{62.1} & \cellcolor[HTML]{C0C0C0}\textbf{77.8} & \cellcolor[HTML]{C0C0C0}\textbf{85.8}\\
        \hline
        LLaMA-3-1B  & Prefix+ICL&21.7&51.6&65.8&44.6&64.6&71.9\\
                    & LoRA &39.3&64.5&78.9&55.4&71.6&79.2\\
        \hline
        LLaMA-3-3B  & Prefix+ICL&21.2&48.9&66.0&46.0&68.9&76.2\\
                    & LoRA &42.1&64.9&78.3&58.7&70.7&79.3\\
        \hline
        LLaMA-3-8B  & Prefix+ICL& 15.9&45.3&62.5&41.9&65.4 & 70.6 \\
                    & LoRA & 35.3 & 65.3 & 72.0 & 54.7 & 74.2 & 83.7\\
        \hline
        LLaMA-3-70B& Prefix+ICL&21.4&57.5&68.5&69.0&89.2&88.3\\
                    & LoRA &32.3&64.8&77.9&57.6&75.8&81.5\\
        \hline
    \end{tabular}
    \label{tab:3_bench_detail}
\end{table}

\begin{table}[h]
    \centering
    \label{tab:3_template_adherence}
    \caption{Template adherence errors across the three best-performing models on 233 test samples.}
    \begin{tabular}{p{6cm}ccc}
    \hline
    \textbf{Evaluation Category} & \textbf{BERT2BERT} & \textbf{LLaMA-3-8B} & \textbf{LLaMA-3-70B} \\
    \hline
    Missing or misspelled headers & 0 & 0 & 0 \\
    \hline 
    Different organ system names & 0 & 14 & 35 \\ \hline
    Inconsistencies in bullet/enumeration formatting
       & 0 & 80 & 61 \\
    \hline 
    Mismatch of mentioned organ systems  & 130 & 136 & 141 \\
    \hdashline
    \quad of which potentially irrelevant & 100 & 113 & 111 \\
    \quad of which potentially relevant & 30 & 23 & 30 \\
    \hline
\end{tabular}
\end{table}

